\documentclass[letterpaper, 10 pt, conference]{ieeeconf}  %

\IEEEoverridecommandlockouts                              %

\makeatletter
\let\NAT@parse\undefined
\makeatother

\usepackage{epsfig}
\usepackage{graphicx}
\usepackage{amsmath}
\usepackage{amssymb}
\usepackage{graphics}
\usepackage{pifont}
\usepackage{color}
\usepackage{booktabs}           %
\usepackage{multirow}
\usepackage{subcaption}
\usepackage{gensymb}
\usepackage{pbox}
\usepackage{xcolor}             %
\usepackage{xspace}             %
\usepackage[sort,square,numbers]{natbib}

\usepackage[breaklinks=true,bookmarks=false]{hyperref}
\hypersetup{
  colorlinks,
  linkcolor={red!50!black},
  citecolor={green!80!black},
  urlcolor={blue!80!black},
}

\graphicspath{{../}{../img}}

\newcommand*{\eg}{e.g.\@\xspace}
\newcommand*{\ie}{i.e.\@\xspace}
\makeatletter
\newcommand*{\etc}{%
    \@ifnextchar{.}%
        {etc}%
        {etc.\@\xspace}%
}
\makeatother

\newcommand{\bx}{\mathbf{x}}

\newcommand{\bw}{\mathbf{w}}

\newcommand{\cL}{\mathcal{L}}
\newcommand{\cG}{\mathcal{G}}

\newcommand{\cS}{\mathcal{S}}
\newcommand{\cT}{\mathcal{T}}

\newcommand{\bel}{\mathrm{Bel}}

\newcommand{\mlane}{\textsc{Lane}}
\newcommand{\mgps}{\textsc{GPS}}
\newcommand{\msign}{\textsc{Sign}}
\newcommand{\lidar}{LiDAR}

\newcommand{\shenlong}[1]{\textcolor{black}{#1}}

\newcommand{\weichiu}[1]{\textcolor{black}{#1}}

\newcommand{\andrei}[1]{\textcolor{black}{#1}}
\newcommand{\andreib}[1]{\andrei{#1}}
\newcommand{\slwang}[1]{\textcolor{black}{#1}}

\title{\LARGE \bf
  Exploiting Sparse Semantic HD Maps for \slwang{Self-Driving Vehicle Localization}
}

\author{Wei-Chiu Ma$^{\ast, 1,2}$, Ignacio Tartavull$^{\ast, 1}$, 
  Ioan Andrei B\^{a}rsan$^{\ast,1,3}$, Shenlong Wang$^{\ast, 1,3}$\\
  Min Bai$^{1,3}$, Gellert Mattyus$^{1}$,
    Namdar Homayounfar$^{1,3}$, Shrinidhi Kowshika Lakshmikanth$^{1}$\\
    Andrei Pokrovsky$^{1}$,
  Raquel Urtasun$^{1,3}$%
\thanks{$^\ast$ Equal contribution}%
\thanks{$^{1}$ Uber Advanced Technologies Group}%
\thanks{$^2$ Department of Electrical Engineering and Computer Science, MIT}%
\thanks{$^{3}$ Department of Computer Science, University of Toronto}%
}

\begin{document}

\maketitle
\thispagestyle{empty}
\pagestyle{empty}

  \vspace{-5cm}
\begin{abstract}
In this paper we propose a novel semantic localization algorithm that exploits
multiple sensors and has precision on the order of a few centimeters. Our
approach does not require detailed knowledge about the appearance of the world, and our
maps \andrei{require orders of magnitude less storage} %
\andrei{than }maps utilized by traditional geometry- and LiDAR intensity-based localizers. This is
important as self-driving cars need to operate in large environments. 
Towards this goal, we formulate the problem in a Bayesian filtering framework,
and exploit lanes, traffic signs, as well as vehicle dynamics to
localize robustly with respect to a sparse semantic map.
We validate the effectiveness of our method on a new highway dataset
consisting of 312km of roads.
Our  experiments show that the proposed approach is able to achieve {0.05m
lateral accuracy and 1.12m longitudinal accuracy} on average while taking up
only {0.3$\%$} of the storage required by previous LiDAR intensity-based approaches.
\end{abstract}

\section{Introduction}

High-definition maps (HD maps) are  a fundamental component of most
self-driving cars, as they contain useful information about the static part of
the environment. The locations of lanes, traffic lights, cross-walks, as well as the 
associated traffic rules are typically encoded in the maps. They encode the
prior knowledge about any scene the autonomous vehicle may encounter.

In order to be able to exploit HD maps, self-driving cars have to localize
themselves with respect to the map. 
The accuracy requirements in localization are very strict and only
a few centimeters of error are tolerable in such safety-critical scenarios.
Over the past few decades, a wide range of localization systems has been
developed.
The Global Positioning System (GPS) exploits triangulation from different
satellites to determine a receiver's position. It is typically affordable, 
but often has several meters of error, particularly in the presence of
skyscrapers and tunnels.
The inertial measurement unit (IMU)  computes the vehicle's  acceleration,
angular rate as well as magnetic field and provides an estimate of its relative
motion, but is subject to drift over time.

To overcome the limitations of GPS and IMU,  place recognition techniques have
been developed. These approaches store what the world looks like either in
terms of geometry (\eg, LiDAR point clouds), visual appearance (\eg, SIFT
features, LiDAR intensity), and/or semantics (\eg, semantic point cloud), and formulate localization as a retrieval task. 
Extensions of classical methods such as iterative closest point (ICP) are
typically employed for geometry-based localization
\cite{yoneda2014lidar,Aghili2016}. 
Unfortunately, geometric approaches suffer in the presence of repetitive
patterns that arise frequently in scenarios such as highways, tunnels, and bridges.
Visual recognition approaches \cite{fabmap} pre-record the scene and  encode
the ``landmark'' visual features. They then perform localization by matching
perceived landmarks to stored ones.
However, they often require capturing the same environment for multiple seasons
and/or times of the day. Recent work \cite{schonberger2018semantic} builds
dense semantic maps of the environment and combines both semantics and geometry
to conduct localization. \shenlong{However, this method requires a large amount
of dense map storage and cannot achieve centimeter-level accuracy.}

\andreib{While place recognition approaches are typically fairly accurate, the
costs associated with ensuring the stored representations are up to date can
often be prohibitive.}
They also require very large storage on board. %
Several approaches have been proposed to provide affordable solutions to localization {by exploiting coarse maps that are freely available on the web \cite{brubaker2013lost,ma2016find}. 
  \andrei{Despite demonstrating promising results, the accuracy of such methods
  is still in the order of a few meters,}
\andrei{which does not meet the requirements of safety-critical applications
such as autonomous driving}.

\begin{figure*}
\centering
    \includegraphics[width=0.95\textwidth]{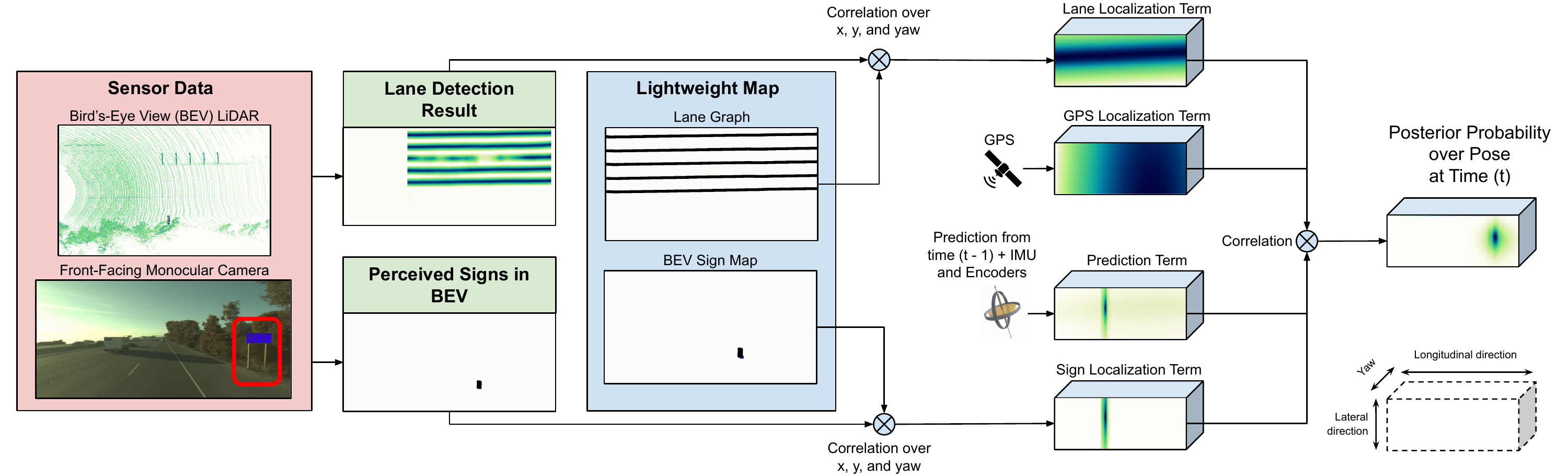}
    \caption{\textbf{System architecture.} Given the camera image and LiDAR
      sweep as input, we first detect lanes in the form of truncated inverse
      distance transform and detect signs as a bird's-eye view (BEV)
      probability map. The detection output is then passed through
      a differentiable rigid transform layer \cite{jaderberg2015spatial} under
      multiple rotational angles. Finally, the inner-product score is measured
      between the inferred semantics and the map. The probability score is
      merged with GPS and vehicle dynamics observations and the inferred pose
      is computed from the posterior using soft-argmax. The camera
    image on the left contains an example of a sign used in localization,
  highlighted with the red box.}
  \vspace{-0.5cm}
\label{fig:network}
\end{figure*}

With these challenges in mind, in this paper we propose a lightweight
localization method that does not require detailed knowledge about the
appearance of the world (\eg, dense geometry or texture).
Instead, we exploit vehicle dynamics as well as a semantic map containing
lane graphs and the locations of traffic signs. \shenlong{Traffic signs provide
information in longitudinal direction, while lanes help avoid lateral drift.}
These cues are complementary to each other and the resulting maps can be stored
in a fraction of the memory necessary for traditional HD maps, which is
important as self-driving cars need to operate in very large environments.
We formulate the localization problem as a Bayes filter,
and demonstrate the effectiveness of our approach on North-American highways,
which are challenging for current place recognition approaches as
repetitive patterns are common and driving speeds are high. Our experiments on
more than 300 km of testing trips showcase that we are able to achieve 0.05m
median lateral accuracy and 1.12m median longitudinal accuracy, while using roughly
three orders of magnitude less storage than previous map-based approaches
(0.55MiB/km$^2$ vs.\ the 1.4GiB/km$^2$ required for dense point clouds).

\begin{figure*}[t]
\centering
\includegraphics[width=0.80\linewidth]{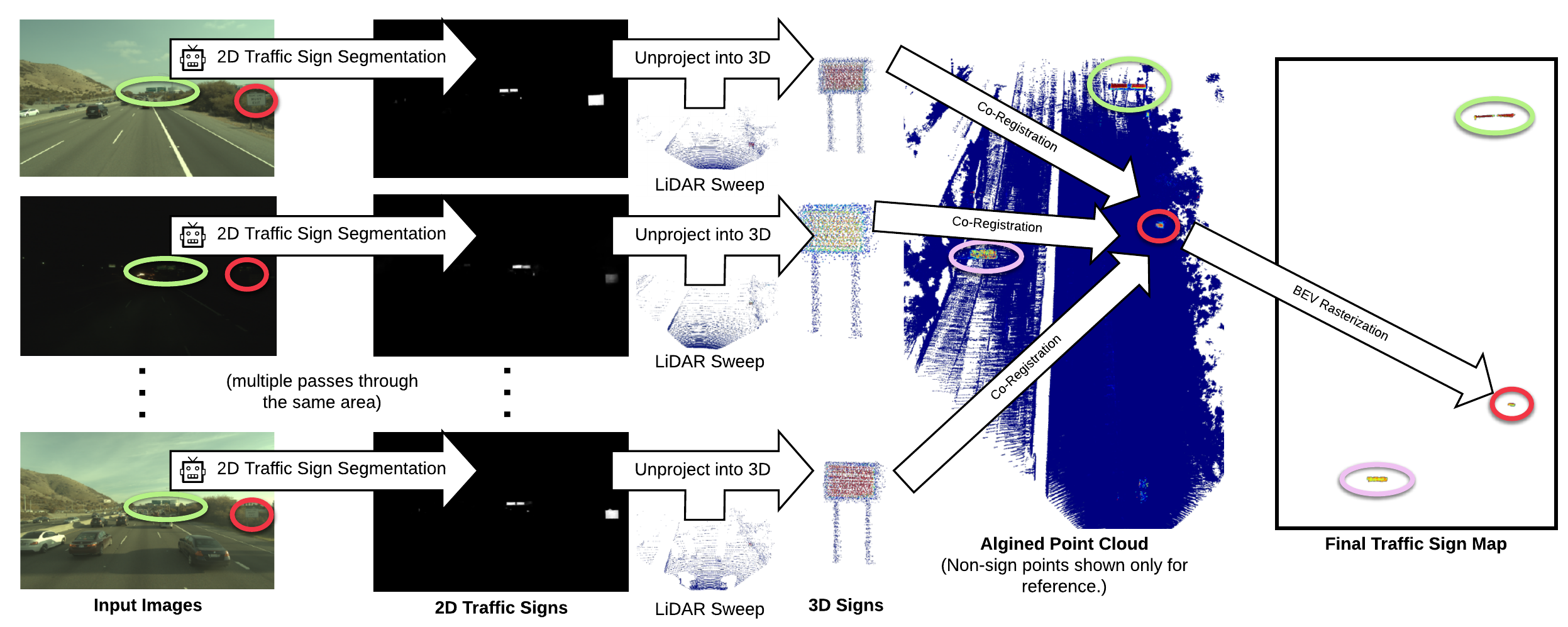}
\caption{\textbf{Traffic sign map building process.} We first detect signs in
2D using semantic segmentation in the camera frame, and then use the LiDAR 
points to localize the signs in 3D. Mapping can aggregate information from
multiple passes through the same area using the ground truth pose information, and can function even in low light, as
highlighted in the middle row, where the signs are correctly segmented even at
night time. This information is used to build the 
traffic sign map in a fully automated way.}
\label{fig:mapping}
\vspace{-0.5cm}
\end{figure*}

\section{Related Work}
\noindent{\bf Place Recognition:}
One of the most prevailing approaches in self-localization is place recognition
\cite{Baatz12,Bansal14,Hays08,Sattler11,snavely,moosmann2013joint,wolcott2014visual,
Arandjelovic2017,Zamir2016}.
By recording the appearance of the world and building a database of it in
advance, the localization task can be formulated as a retrieval problem. At
test time, the system simply searches for the most
similar scene and retrieves its pose. As most of the features used to describe the scene 
(e.g., 3D line segments \cite{Bansal14} or 3D point clouds
\cite{snavely,moosmann2013joint,wolcott2014visual}), 
are highly correlated with the appearance of the world, one  needs to update
the database frequently. With this problem in mind, \cite{fabmap,
nelson2015dusk,linegar2015work} proposed an image-based localization technique
that is to some degree invariant to appearance changes. More recently,
\cite{kendall2015posenet} designed a CNN to directly estimate the pose of the
camera. While their method is robust to illumination changes and weather, they
still require training data for each scene, limiting its scalability.

\vspace{0.1cm}
\noindent{\bf Geometry-based Localization:} Perspective-n-Point (PnP)
approaches have been used for localization. The idea is to extract local
features from images and find  correspondences with the pre-stored geo-registered point sets. 
For instance, \cite{jamieshotton} utilized random forests to find
correspondence between  RGBD images and pre-constructed 3D indoor geometry.
Li et al.~\cite{snavely} pre-stored point clouds along with  SIFT features for
this task, while
Liu et al.~\cite{hongdongli} proposed to use branch-and-bound to solve the exact   2D-3D
registration. However, these approaches require computing a 3D reconstruction 
of the scene in advance, and do not work well in scenarios with repetitive 
geometric structures.}%

\vspace{0.1cm}
\noindent{\bf Simultaneous Localization and Mapping:} Given a sequence of
images, point clouds, or depth images, SLAM approaches \cite{lsdslam, Mur-Artal2017,
jizhang} estimate relative camera poses between consecutive frames through
feature matching and joint pose estimation. 
Accumulated errors  make the localization gradually drift as the robot moves.
In indoor or urban scenes, loop closure has been used to fix the accumulated
errors. However, unlike indoor or urban scenarios, on highways trajectories are 
unlikely to be closed, which makes drift a much more challenging problem to
overcome.

\vspace{0.1cm}
\noindent{\bf Lightweight Localization:} There is a growing interest in 
developing affordable localization techniques. Given an initial estimate of the vehicle position, \cite{Floros13} exploited ego-trajectory to self-localize within a small region.
Brubaker et al.~\cite{brubaker2013lost} developed a technique that can be applied at city
scale, without any prior knowledge about the vehicle location.
Ma et al.~\cite{ma2016find} incorporated other visual cues, such as the position of the
sun and the road type to further improve the results. These works are appealing since they only require a cartographic map. However, the localization accuracy is strongly limited by the performance of odometry. The semantic cues are only used to resolve ambiguous modes and speed up the inference procedure. Second, the computational complexity is a function of the uncertainty in the map, which remains fairly large when dealing with maps that have repetitive structures.

\vspace{0.1cm}
\noindent{\bf High-precision Map-based Localization:} The proposed work belongs
to the category of the high-precision map-based localization
\cite{levinson2007map, levinson2010robust, wolcott2015fast, wolcott2014visual,
schreiber2013laneloc, yoneda2014lidar, ziegler2014video, deep-gil}.
\andrei{The use of maps has been shown to not only provide strong cues for
various tasks in computer vision and robotics such as scene understanding
\cite{WangCVPR15}, vehicle detection \cite{matzen2013nyc3dcars}, and
localization~\cite{brubaker2013lost,WangICCV15,ma2016find}, but also enables 
the creation of large-scale datasets with little human
effort~\cite{wegner2016cataloging,wang2016torontocity}.}
The general idea is
to build a centimeter-level high-definition 3D map offline a priori, by
stitching sensor input
using a high-precision differential GNSS system and offline SLAM.  Early approaches utilize LiDAR sensors to build maps \cite{levinson2007map}. Uncertainty in intensity changes have been handled through building 
probabilistic prior map \cite{levinson2010robust, wolcott2015fast}.  
In the online stage, the position is determined by matching the sensor reading 
to the prior map. For instance,~\cite{levinson2007map, levinson2010robust,
wolcott2015fast} utilized the perceived LiDAR intensity to conduct matching.
Yoneda et al.~\cite{yoneda2014lidar} proposed to align online LiDAR sweeps
against an existing 3D prior map using ICP,
\cite{ziegler2014video,wolcott2014visual} utilized visual cues from cameras to
localize the self-driving vehicles, and
\cite{deep-gil} use a fully convolutional neural network to learn the task of
online-to-map matching in order to improve robustness to dynamic objects and
eliminate the need for \lidar{} intensity calibration.

\vspace{0.1cm}
\noindent{\bf Semantic Localization:} Schreiber et al. \cite{schreiber2013laneloc} 
proposed to use lanes as localization cues. Towards this goal, they  manually annotated lane markings over the LiDAR intensity map. 
The lane markings are then detected online using a stereo camera, and matched
against the ones in the map. 
\shenlong{Welzel et al.~\cite{welzel2015improving} and Qu et al.~\cite{qu2015vehicle} utilize traffic signs to 
assist image-based vehicle localization. Specifically, traffic signs are
detected from images and matched against a geo-referenced sign database, after
which local bundle adjustment is conducted to estimate a fine-grained pose.}
More recently,~\cite{schonberger2018semantic} built dense semantic maps using image segmentation and conducted localization by matching both semantic and geometric cues. 
In contrast, the maps used in our approach only need to contain the lane graphs 
and the inferred sign map, the latter of which is computed without a human in
the loop, while also only requiring a fraction of the storage used by dense
maps.

\def\arraystretch{0.925}%
\section{Lightweight HD Mapping}

In order to conduct efficient and accurate localization, a compressed yet
informative representation of the world needs to be constructed.
Ideally our HD maps should be easy to (automatically) build and maintain at scale,
while also enabling real-time broadcasting of map changes between a central server 
and the fleet.  
This places stringent storage requirements that traditional dense HD maps fail
to satisfy.
In this paper, we tackle these challenges by building sparse HD maps
containing just the lane graph and the locations of traffic signs.
These modalities provide complementary semantic cues for localization.  Importantly, the storage needs for our maps are  
three orders of magnitude smaller than traditional \lidar{} intensity maps~\cite{levinson2010robust, wolcott2015fast, levinson2007map, deep-gil} or geometric maps \cite{yoneda2014lidar}. 

\paragraph{Lane Graph} Most roads have visually distinctive  lanes determining
the expected trajectory of vehicles, compliant with the traffic rules.  
Most self driving cars store this prior knowledge as lane graphs $\mathcal{L}$.
A lane graph   is a structured representation of the road network defined as
a set of polygonal chains (polylines), each of which represents a lane boundary.   
We refer the reader to Fig.~\ref{fig:network} for an illustration of a lane graph. 
Lane graphs provide useful cues for localization, particularly in the lateral  position and the heading of the vehicle.

 \paragraph{Traffic Signs} Traffic signs are common semantic landmarks that are
 sparsely yet systematically present  in cities, rural areas, and highways. Their presence  provides useful cues that can be employed for accurate longitudinal  localization.
 In this paper we build sparse HD maps containing traffic signs automatically. 
 Towards this goal, we exploit multiple passes of our vehicles over the same region and identify the signs by exploiting image-based 
 semantic segmentation followed by 3D sign localization using LiDAR via \shenlong{inverse-projection from pixel to 3D space}. 
 A consistent coordinate system over the multiple passes is obtained by means of  offline multi-sensor SLAM.
Note that in our map we only store points that are estimated to be traffic
signs above a certain confidence level. 
After that, we rasterize the sparse points to create the traffic sign presence
probability map $\mathcal{T}$ in bird's-eye view (BEV) at 5cm per pixel. 
This  is a very sparse representation containing all the traffic signs. 
The full process is conducted without any human intervention. %
Fig.~\ref{fig:mapping} depicts the traffic sign map building process
and an example of its output.

\section{Localization as Bayes Inference with Deep Semantics}

In this paper we propose a novel localization system that exploits vehicle
dynamics as well as a semantic map containing both a lane graph and the
locations of traffic signs.
These cues are complementary to each other and the resulting maps can be stored
in a fraction of the memory necessary for traditional dense HD maps.  
We formulate the localization problem as a histogram filter taking as input the
structured outputs of our sign and lane detection neural networks, as well as
GPS, IMU, and wheel odometry information, and
outputting a probability histogram over the vehicle's pose, expressed in world
coordinates.

\vspace{-0.2cm}
\subsection{Probabilistic Pose Filter Formulation}
\setcounter{paragraph}{0}

Our localization system exploits a wide variety of sensors: GPS, IMU, wheel
encoders, LiDAR, and cameras. These sensors are available in most
self-driving vehicles. The GPS provides a coarse location with
several meters accuracy; an IMU captures vehicle dynamic measurements; the
wheel encoders measure the total travel distance;
the LiDAR accurately perceives the geometry of the surrounding area through a
sparse point cloud; images capture dense and rich appearance information.
We assume our sensors are calibrated and neglect the effects of suspension, %
unbalanced tires, and vibration. As shown in our experiments, the influence of
these factors is negligible and other aspects such as sloped roads (\eg, on
highway ramps) do not have an impact on our localizer.
Therefore, the vehicle's pose can be parametrized with only three degrees of
freedom (instead of six) consisting of a 2D translation and a heading angle
\shenlong{w.r.t.\ the map coordinate's origin}, \ie$\mathbf{x} = \{\mathbf{t}, \theta \}$, where $\mathbf{t} \in \mathbb{R}^2$ and $\theta \in (-\pi, \pi]$, since the heading is parallel to the ground plane.

Following similar concepts to~\cite{deep-gil}, we factorize the posterior
distribution over the vehicle's pose into components corresponding to each
modality, as shown in Equation~\eqref{eq:inference}.

Let $\mathcal{G}_t$  be the GPS readings at time $t$ and let $\mathcal{L}$ and $\mathcal{T}$ represent the lane graph and traffic sign maps respectively. 
We compute an estimate of the vehicle dynamics 
$\mathcal{X}_t$
from both IMU  and the wheel encoders smoothed through an extended Kalman
filter, which is updated at 100Hz.

The localization task is formulated as a histogram filter aiming to maximize
the agreement between the observed and mapped lane graphs and traffic signs
while respecting vehicle dynamics:
\begin{align}
\mathrm{Bel}_t(\mathbf{x}) = \eta \cdot &P_{\mlane}(\cS_t | \bx, \cL;
\bw_\mlane)  P_{\msign}(\cS_t | \bx, \cT; \bw_\msign) \nonumber \\
&P_{\mgps}(\cG_t | \bx) \mathrm{Bel}_{t|t-1} (\bx | \mathcal{X}_t),
 \label{eq:inference}
\end{align}
where $\bel_t(\mathbf{x})$ is the posterior probability of the vehicle pose
at time $t$; $\eta$ is a normalizing factor to ensure sum of all probability is
equal to one; $\bw_\mlane$ and $\bw_\msign$ are sets of learnable parameters,
and
$\mathcal{S}_t = (\mathcal{I}_t, \mathcal{C}_t)$ is a sensory measurement
tuple composed from LiDAR $\mathcal{I}_t$ and camera $\mathcal{C}_t$,
Note that
by recursively solving Eq.~\eqref{eq:inference}, we can localize the vehicle at
every  step with an uncertainty measure that could be propagated to the next
step.  We now describe each energy term and our inference algorithm in more
detail.

\begin{figure*}[]
\centering
  \includegraphics[width=0.95\textwidth]{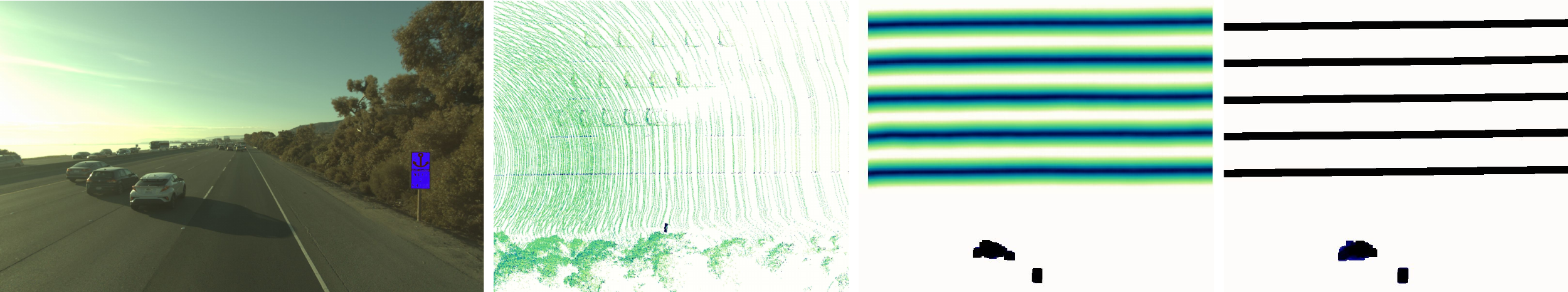}

\caption{
    \textbf{Dataset sample and inference results.} Our system detects signs in
    the camera images (note the blue rectangle on the right side of the first image)
    and projects
    the sign's points in a top-down view using LiDAR (second image). 
    It uses this result in
    conjunction with the lane detection result (third image) to localize
    against a lightweight map consisting of just signs and lane boundaries
    (fourth image).}
\vspace{-0.5cm}
\label{fig:sample}
\end{figure*}

\paragraph{Lane Observation Model}
We define our matching energy  to encode the agreement between the lane observation from the sensory input and the map. 
Our probability is  computed by  a normalized matching score function that utilizes the existing lane graph and compares it to  detected lanes. 
To detect lanes we  exploit a state-of-the-art real-time multi-sensor convolutional network~\cite{min2018deep}.
The input of the network is a front-view camera image and raw LiDAR intensity
measurement projected onto BEV. The output of the network is
the inverse truncated distance function to the lane graph in the overhead view.
Specifically, each pixel in the overhead view encodes the Euclidean distance to the
closest lane marking, up to a truncation threshold of 1m.
We refer the reader to Fig.~\ref{fig:sample} for an illustration of the neural
network's input and output. 

To compute the probability, we first orthographically project the lane graph
$\cL$ onto overhead view such that the lane detection output and the map are
under the same coordinate system. The overhead view of the lane graph is also
represented using an truncated inverse distance function.
Given a vehicle pose hypothesis $\mathbf{x}$, we rotate and translate the lane detection prediction accordingly and compute its matching score against the lane graph map.  The matching score is an inner product between the lane detection and the lane graph map
\begin{align} \label{eq:lane}
P_{\mlane} \propto s \left(\pi\left(f_\mlane(\cS; \bw_\mlane), \bx\right), \cL \right),
\end{align}
where $f_\mlane$ is the deep lane detection network and $\bw_\mlane$ are the
network's parameters. $\pi$ is a 2D rigid transform function to transform the online
lane detection to the map's coordinate system  given
a pose hypothesis $\bx$; $s(\cdot, \cdot)$ is a cross-correlation operation between two images.

\paragraph{Traffic Sign Observation Model}
This model encodes the consistency between perceived online traffic signs and the  map. Specifically, we  run an image-based semantic segmentation algorithm that performs dense semantic labeling of traffic signs. We adopt the state-of-the-art PSPnet structure \cite{zhao2017pyramid} to our task. The encoder architecture is a ResNet50 backbone and the decoder is a pyramid spatial pooling network. Two additional convolutional layers are added in the decoder stage to further boost performance.  The model is jointly trained with the instance segmentation loss following~\cite{bai2017deep}. 
Fig.~\ref{fig:sample} depicts examples of the network's input and output.
The estimated image-based traffic sign probabilities are converted onto the
overhead view to form our online traffic sign probability map. This is achieved
by associating each LiDAR  with a pixel in the image by projection. We then
read the softmax probability of the pixel's segmentation  as our estimate.
Only high-confident traffic sign pixels are unprojected to 3D and rasterized in
BEV. Given a pose proposal $\bx$, we define the sign matching probability
analogously to the lane matching one as
\begin{align} \label{eq:sign}
P_{\msign} \propto s \left(\pi\left(f_\msign(\cS; \bw_\msign), \bx\right), \cT \right),
\end{align}
where $f_\msign$ is the sign segmentation network and $\bw_\msign$ are the networks' parameters.
Both the perceived signs, as well as the map they are matched against are
encoded as pixel-wise occupancy probabilities.

\paragraph{GPS Observation Model}
This term  encodes the likelihood of the GPS sensory observation $\cG$ at a given vehicle pose $\mathbf{x}$:
\begin{align} \label{eq:gps}
P_{\mgps} \propto \exp\left( - \frac{(g_x - x)^2 + (g_y - y)^2}{\sigma_\mgps^2}\right),
\end{align}
where $[g_x, g_y]^T = T \ \cdot \ \cG$ represents a GPS point location in the
\andrei{coordinate frame of the map against which we are localizing.}
$T$ is the given rigid transform between the Universal Transverse Mercator (UTM)
coordinates and the map coordinates and $\cG$ is the GPS observation expressed
in UTM coordinates.

\paragraph{Dynamics Model}
This term encourages consistency between the pose proposal $\mathbf{x}$ and the
vehicle dynamics estimation, given the previous  vehicle pose distribution. The
\andrei{pose at the current timestamp depends on previous pose and the
vehicle motion}. Given an observation of the \andrei{vehicle} motion $\mathcal{X}_t$,
the motion model is computed by marginalizing out the previous pose:
 \begin{equation}
  \label{eq:motion-model}
  \mathrm{Bel}_{t|t-1} (\bx | \mathcal{X}_t) = \sum_{\bx_{t-1} \in
  \mathcal{R}_{t-1}} P(\bx | \mathcal{X}_t, \bx_{t-1}) \mathrm{Bel}_{t-1}(\bx_{t-1})
\end{equation}
where the likelihood is a Gaussian probability model
\begin{equation} \label{eq:dynamics-energy}
P(\bx | \mathcal{X}_t, \bx_{t-1})  \propto \mathcal{N}(\left(\bx \ominus (\bx_{t-1} \oplus \mathcal{X}_t)\right), \Sigma)
\end{equation}
with $\Sigma$ the covariance matrix  and $\mathcal{R}_{t-1}$ is the search
space for $\bx_{t-1}$. In practice, we only need $\mathcal{R}_{t-1}$ to be
a small local region centered at $\bx_{t-1}^\ast$ given the
fact the rest of the pose space has negligible probability.
Note that $\oplus$, $\ominus$ are the standard 2D pose composition and
inverse pose composition operators described by Kummerle et
al.~\cite{kummerle2009measuring}.

\vspace{-0.2cm}
\subsection{Efficient Inference}
\label{sec:inference}
\setcounter{paragraph}{0}

\paragraph{Discretization} The inference defined in Eq.~(\ref{eq:inference})  is intractable.  Following \cite{deep-gil,levinson2010robust} we tackle this problem using a histogram filter. We discretize the full continuous search space over 
$\bx = \{ x, y, \theta \}$  into a search grid,  each with associated 
posterior $\bel(\bx)$. 
We restrict the search space to  a small local region at each time. This is a reasonable assumption given the constraints of the vehicles dynamics at a limited time interval. 

\paragraph{Accelerating Correlation}  We now discuss the computation required
for each term. We utilize  efficient convolution-based exhaustive search to
compute the lane and traffic sign probability model. In particular, enumerating
the full translational search range with inner product is equivalent to
a correlation filter with a large kernel (which is the online sign/lane
observation). Motivated by the fact that the kernel is very large, FFT-conv is
used to accelerate the computation speed by a factor of $20$ over
the state-of-the-art GEMM-based spatial GPU correlation implementations~\cite{deep-gil}.

\paragraph{Robust Point Estimation}  Unlike the MAP-inference which simply 
takes the configuration which maximizes the posterior belief, we adopt a center-of-mass based soft-argmax~\cite{levinson2010robust} to better incorporate the uncertainty  of our model and encourage smoothness in our localization. We thus define
\begin{equation}
\label{eq:soft-argmax}
\bx_t^\ast= \frac{\sum_\bx \mathrm{Bel}_t(\bx)^\alpha \cdot \bx}{\sum_\bx
\mathrm{Bel}_t(\bx)^\alpha},
\end{equation}
where $\alpha \geq 1$ is a temperature hyper-parameter. This gives us an
estimate that takes the uncertainty of the prediction into account.

\vspace{-0.2cm}
\subsection{Learning}
\setcounter{paragraph}{0}
Both the lane detection network and the traffic sign segmentation network are trained through back-propagation separately using ground-truth annotated data. The lane detection is trained with a regression loss that measures 
the $\ell_2$ distance between the predicted inverse truncated distance
transform and the ground-truth one~\cite{min2018deep}.
The semantic segmentation network is trained with
cross-entropy~\cite{bai2017deep}. Hyper-parameters for the Bayes filter (\eg,
$\sigma^2_{\text{GPS}}$, softmax temperature $\alpha$, etc.) are
searched through cross-validation.

\section{Experiments}
We validate the effectiveness of our localization system on a highway dataset
of \weichiu{312km}. We evaluate our model in terms of its localization
accuracy
and runtime.

\vspace{-0.1cm}
\subsection{Dataset}
Our goal is to perform fine-grained localization on highways. Unfortunately,
there is no publicly available dataset that provides ground truth localization
at the centimeter-level precision required for safe autonomous driving.
We therefore collected a dataset of highways by driving over 300km in North
America at different times of the year, covering over 100km of roads.
The dataset encompasses 64-line LiDAR sweeps and images from a front-facing global shutter camera with
a resolution of $1900\times1280$, both captured at 10Hz, as well as IMU and GPS
sensory data and the lane graphs. \slwang{The extrinsic calibration between the
camera and LiDAR is conducted} \andrei{using a set of calibration
targets~\cite{hartley2003multiple}.}
\slwang{The ground truth 3D localization is estimated by a high-precision ICP-based offline Graph-SLAM using high-definition pre-scanned scene geometry.}
Fig.~\ref{fig:sample} shows \andrei{a sample from our dataset
together with the inferred and ground truth lane graphs}. %

Our dataset is partitioned into `snippets', each consisting of roughly 2km of
driving.  The training, validation, and test splits are conducted at the
snippet level, where training snippets are used for map building and training
the lane
detection network, and validation snippets are used for hyper-parameter tuning.
The test snippets are used to compute the final metrics.
An additional 5,000 images have been annotated with pixel-wise traffic sign
labels which are used for training the sign segmentation network.

\subsection{Implementation Details}

\noindent{{\bf Network Training:}} 
To train the lane detection network, we uniformly sample 50K frames from the training region based on their
geographic coordinates to train the network. The ground truth can be
automatically generated given the vehicle pose and the lane graph. We use
a mini-batch size of 16 and employ Adam~\cite{kingma2014adam} as the optimizer. We set the learning rate
to $10^{-4}$. The network was trained completely from scratch with Gaussian
initialization and converged roughly after 10 epochs. We visualize some results
in Fig.~\ref{fig:sample}.

We train our traffic sign segmentation network separately over four GPUs with
a total mini-batch size of 8. 
Synchronized batch normalization is utilized for multi-GPU batch normalization.
The learning rate is set to be $10^{-4}$ and the network is trained from
scratch. The backbone of the model is fine-tuned from a DeepLab v2 network
pre-trained over the Pascal VOC dataset.

\noindent{{\bf Hyper-parameter Search:}} 
We choose the hyper-parameters through grid search over a mini-validation
dataset consists of 20 snippets of 2km driving. The hyper-parameters include
the temperatures of the final pose soft-argmax, the lane
probability softmax, and the sign probability softmax, as well as the 
observation noise parameters for GPS and the dynamics.
The best configuration is chosen by the failure rate metric. In the
context of hyperparameter search, the failure rate is a snippet-level metric
which counts a test snippet as failed if the total error becomes greater than
1m at any point. We therefore picked the hyperparameter configuration which
minimized this metric on our validation set, and kept it fixed at test time.
As noted in Sec.~\ref{sec:inference}, we restrict our search range to a small area centered at the dead reckoning pose and neglect the probability outside the region. 
We notice in practice that thanks to the consistent presence of the lanes in self-driving scenarios, 
there is less uncertainty along the lateral direction than along the longitudinal. 
The presence of traffic signs helps reduce uncertainty along the longitudinal direction, 
but signs could be as sparse as every 1km, during which INS drift could be as large as 7 meters. Based on this observation and with the potential drift in mind, we choose a very conservative search range 
$B = B_x \times B_y \times B_\theta = \left[ -0.75m, 0.75m \right] \times
\left[ -7.5m, 7.5m \right] \times \left[ -2^\circ, 2^\circ \right]$ at
a spatial resolution of 5cm and an angular resolution of $1^\circ$.

\subsection{Localization}
\noindent{{\bf Metrics:}}
We adopt several key metrics to measure the localization performance of the
algorithms evaluated in this Section.

In order to safely drive from a certain point to another without any human
intervention, an autonomous vehicle must be aware of where it is w.r.t.\ the 
map. Lateral error and longitudinal error have different meanings for
self-driving
since a small lateral error could result in localizing in the wrong lane, while
ambiguities about the longitudinal position of the vehicle are more tolerable.
As localization is the first stage in self-driving pipeline, it it critical
that it is robust enough with a very small failure rate;
therefore, understanding worst-case performance is critical.

Moreover, localization results should reflect the vehicle dynamics as well, 
which ensures the smoothness of decision making, since sudden jumps in
localization might cause downstream components to fail. To this end, we also
measure the prediction smoothness of our methods.
We define smoothness as the
difference between the temporal gradient of the ground truth pose and that of
the predicted poe. We estimate the gradients using first-order finite
differences, i.e., by simply taking the differences between poses at times
$(t)$ and $(t-1)$. As such, we define smoothness as
\begin{equation}
  s = \frac{1}{T} \sum_{t=1}^T \left\| (\bx_t^\ast - \bx_{t-1}^\ast) - (\bx_t^{GT} - \bx_{t-1}^{GT}) \right\|^2.
\end{equation}

\vspace{0.1cm}
\noindent{{\bf Baselines:}}
We compare our results with two baselines: dynamics and dynamics+GPS. The first
baseline builds on top of the dynamics of the vehicle. It takes as input the
IMU data and wheel odometry, and use the measurements to extrapolate the
vehicle's motion. The second baseline employs histogram filters to fuse
information between IMU readings and GPS sensory input, which combines motion
and absolute position cues.

\noindent{{\bf Quantitative Analysis:}} 
As shown in Tab.~\ref{tab:comparison} and~\ref{tab:comparison-smooth}, our method significantly outperforms the
baselines across all metrics. To be more specific, our model has a median
longitudinal error of 1.12m and a median lateral error of 0.06m;  both are much
smaller than other competing methods, with lateral error one order of magnitude
lower. We notice that our method greatly improve the performance over the worst
case scenario in terms of both longitudinal error, lateral error, and
smoothness.

\setlength\tabcolsep{4pt}   %
\begin{table}
\center
\caption{Quantitative results on localization accuracy. Here, `Ours' refers to
the model proposed in this paper using dynamics, GPS, lanes, and signs, in
a probabilistic framework%
.}
\label{tab:comparison}
\vspace{-0.1cm}
\begin{tabular}{lrrrrrr}
\toprule
\multirow{3}{*}{Methods} & \multicolumn{3}{c}{Longitudinal Error (m)}
                         & \multicolumn{3}{c}{Lateral Error (m)}    \\
                         \cmidrule{2-7} 
                         & Median       & 95\%         & 99\% & Median
                         & 95\%         & 99\%         \\ \midrule
Dynamics             & 24.85         & 128.21        & 310.50        & 114.46
                     & 779.33        & 784.22       \\
GPS                  & 1.16 & 5.78          & 6.76          & 1.25
                     & 8.56          & 9.44         \\
INS                      & 1.59          & 6.89          & 13.62         & 2.34
                         & 11.02         & 42.34         \\
                         Ours                 & \textbf{1.12} & \textbf{3.55}
                                              & \textbf{5.92} & \textbf{0.05}
                                              & \textbf{0.18} & \textbf{0.23} \\
                         \bottomrule
\end{tabular}
\vspace{-0.1cm}
\end{table}
\setlength\tabcolsep{6pt}

\begin{table}
\centering
\caption{Quantitative results on smoothness}
\vspace{-0.1cm}
\begin{tabular}{lrrrr}
\toprule
\multirow{2}{*}{Method Name} & \multicolumn{4}{c}{Smoothness}
\\ \cmidrule{2-5} 
                             & Mean                  & 95\%         & 99\%
                             & Max          \\ \midrule
Dynamics                & 0.2                   & 0.4          & 0.6
                        & 1.2          \\
GPS                   & \textbf{0.1}          & 0.2          & 0.3          & 8.5          \\
INS                          & \textbf{0.1}          & \textbf{0.1} & \textbf{0.2} & 3.7          \\
Ours                         & \textbf{0.1}          & 0.2          & 0.3
                             & \textbf{0.9} \\ \bottomrule
\end{tabular}
\vspace{-0.5cm}
\label{tab:comparison-smooth}
\end{table}

\begin{table*}[]
\caption{Ablation studies on the impact of each system component}
\vspace{-0.2cm}
\centering
\begin{tabular}{lrrrrrrrrr}
\toprule
\multirow{4}{*}{Method} & \multicolumn{3}{c}{\multirow{2}{*}{Properties}}
                        & \multicolumn{6}{c}{Travelling Dist $=2km$}
                        \\ \cmidrule{5-10} 
                             & \multicolumn{3}{c}{}
                             & \multicolumn{3}{c}{Longitudinal Error (m)}
                             & \multicolumn{3}{c}{Lateral Error (m)}
                             \\ \cmidrule{2-10}
                             & Lane              & GPS              & Sign
                             & Median          & 95\%         & 99\%
                             & Median          & 95\%         & 99\%         \\
                             \midrule
Lane                         & yes               & no               & no               
                             & 13.45         & 37.86         & 51.59         
                             & 0.20          & 1.08          & 1.59          \\ 
Lane+GPS                     & yes               & yes              & no               
                             & 1.53          & 5.95          & 6.27          
                             & 0.06          & 0.24 & 0.43 \\ 
Lane+Sign                    & yes               & no               & yes              
                             & 6.23         & 31.98         & 51.70         
                             & 0.10          & 0.85          & 1.41          \\
                             \midrule
All                          & yes               & yes              & yes              
                             & \textbf{1.12} & \textbf{3.55} & \textbf{5.92} 
                             & \textbf{0.05} & \textbf{0.18} & \textbf{0.23}
                             \\ \bottomrule
\end{tabular}
\label{tab:ablation}
\vspace{-0.2cm}
\end{table*}

\begin{table*}[]
\centering
\caption{Ablation studies on inference settings with full observations
(Lane+GPS+Sign)}
\vspace{-0.2cm}
\begin{tabular}{lrrrrrrrrrrr}
\toprule
\multirow{4}{*}{Inference} & \multicolumn{6}{c}{Travelling Dist $=2km$}
                           & \phantom{}
                           & \multicolumn{4}{c}{\multirow{2}{*}{Smoothness}}
                           \\ \cmidrule{2-7}
                                        & \multicolumn{3}{c}{Longitudinal Error (m)}
                                        & \multicolumn{3}{c}{Lateral Error (m)}
                                        & \multicolumn{4}{c}{} \\
                                        \cmidrule{2-12}
                                        & Median       & 95\%         & 99\%
                                        & Median       & 95\%         & 99\%
                                        &
                                        & Mean                  & 95\%
                                        & 99\%         & Max          \\
                                        \midrule
Deterministic                                      
  & 1.29 & 3.65 & \textbf{5.16}          
  & 0.08          & 0.26          & 0.50
  &
  & 0.11          & \textbf{0.19} & 1.78          & 5.27          \\
Probabilistic                                     
  & \textbf{1.12} & \textbf{3.55} & 5.92 
  & \textbf{0.05} & \textbf{0.18} & \textbf{0.23}
  &
  & \textbf{0.07}          & \textbf{0.19} & \textbf{0.24} & \textbf{0.98} \\
  \bottomrule
\end{tabular}
\vspace{-0.1cm}
\label{tab:ablation-det-prob}
\end{table*}

\begin{figure*}[]
\centering

  \includegraphics[width=0.85\textwidth]{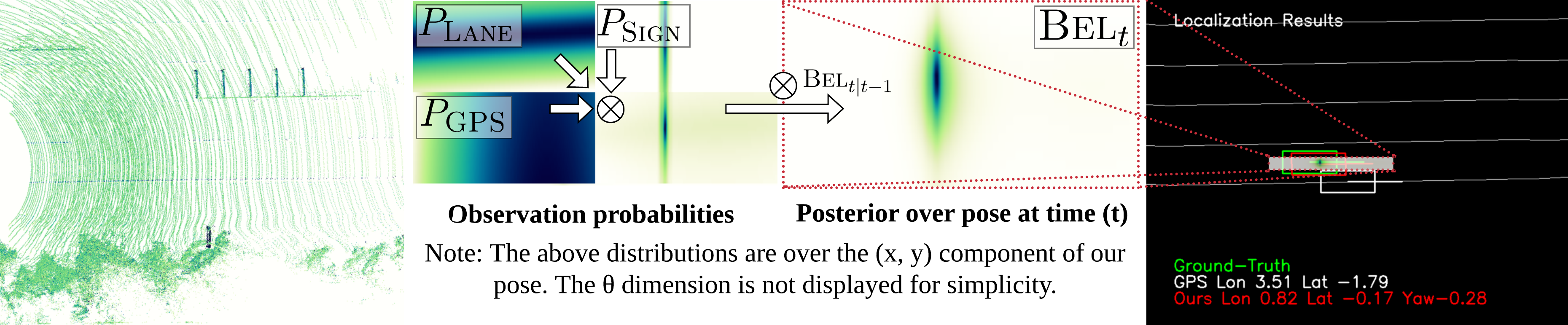}
  \caption{\textbf{Qualitative results.}
    A bird's-eye view of the last five LiDAR sweeps (left), which are used for
         the lane detection, together with the observation probabilities and
         the posterior (middle), followed by a comparison between the localization
         result, the ground truth pose, and GPS (right). The (x, y)-resolution
         of each probability distribution is 1.5m laterally (vertical) and 15m
         longitudinally (horizontal).
}
         \vspace{-0.5cm}
\label{fig:localization}
\end{figure*}

\vspace{0.1cm}
\noindent{{\bf Qualitative Results:}} 
We show the localization results of our system as well as those of the
baselines in Fig. \ref{fig:localization}. \shenlong{Through lane observations,
our model is able to consistently achieve centimeter-level lateral localization
accuracy. When \andrei{signs are visible,} 
the traffic sign model helps push the prediction towards
the location where the observation and map have agreement, bringing 
the pose estimate to the correct longitudinal position. 
In contrast, GPS tends to produce noisy results, but helps substantially
improve worst-case performance.}%

\noindent{{\bf Runtime Analysis:}} 
To further demonstrate that our localization system is of practical usage, we
benchmark the runtime of each component in the model during inference using an
NVIDIA GTX 1080 GPU.
A single step of our inference takes 153ms in total on average, with 
32ms on lane detection, 110ms on semantic segmentation 
and 11ms on matching, which is roughly 7 fps. %
We note that the real-time performance is made possible largely with the help of FFT convolutions.

\noindent{\bf Map Storage Analysis:}
\weichiu{We compare the size of our HD map against other commonly used
representations: LiDAR intensity map and 3D point cloud map. For a fair
comparison, we store all data in a lossless manner and measure the storage
requirements. While the LiDAR intensity and 3D point cloud maps consume 177 MiB/km$^2$ and 1,447 MiB/km$^2$ respectively, our HD map only requires \textbf{0.55 MiB} per square kilometer. This is only $0.3\%$ of the size of LiDAR intensity map and $0.03\%$ of that of 3D point cloud map.}

\noindent{{\bf Ablation Study:}} 
To better understand the contribution of each component of our model, we
respectively compute the longitudinal and lateral error under diverse settings.
As shown in Tab.~\ref{tab:ablation}, each term (GPS, lane, sign) has a positive
contribution to the localization performance.
Specifically, the lane observation model greatly increases lateral accuracy, 
while sign observations increase longitudinal accuracy.
We also compare our probabilistic histogram filter formulation with a deterministic 
model. Compared against our histogram filter approach, the non-probabilistic 
one performs a weighted average between each observation without carrying over
the previous step's uncertainty. 
As shown in Tab.~\ref{tab:ablation-det-prob}, by combining all the observation models,
the non-probabilistic model can achieve reasonable performance but still
remains less accurate than the probabilistic formulation. Moreover, due to the fact 
that no uncertainty history is carried over, prediction smoothness over time is
not guaranteed. 

\section{Conclusion}

In this paper we proposed a robust localization system capable of localizing
an autonomous vehicle
against a map requiring roughly three orders of magnitude less storage than
traditional methods. This has the potential to substantially
improve the scalability of self-driving technologies by reducing storage
costs, while also enabling map updates to be delivered to vehicles at
vastly reduced costs.

We approached the task by identifying two sets of complementary cues capable of
disambiguating the lateral and longitudinal position of the vehicle: lane
boundaries and traffic signs. We integrated these
cues into a pipeline alongside GPS, IMU, and wheel encoders, and showed that
the system is able to run in real time at roughly 7Hz on a single GPU. We
demonstrated the efficacy of our method on a large-scale highway
dataset consisting of over 300km of driving, showing that it can achieve
\andrei{the localization accuracy requirements of self-driving cars}, while
using much less storage.

{\small
    \bibliographystyle{ieee}
    \bibliography{egbib}

\begin{thebibliography}{10}\itemsep=-1pt

\bibitem{Aghili2016}
F.~Aghili and C.~Y. Su.
\newblock {Robust relative navigation by integration of ICP and adaptive Kalman
  filter using laser scanner and IMU}.
\newblock {\em TMECH}, 2016.

\bibitem{Arandjelovic2017}
R.~Arandjelovic, P.~Gronat, A.~Torii, T.~Pajdla, and J.~Sivic.
\newblock {NetVLAD: CNN architecture for weakly supervised place recognition}.
\newblock {\em IEEE TPAMI}, 2017.

\bibitem{Baatz12}
G.~Baatz, K.~K{\"o}ser, D.~Chen, R.~Grzeszczuk, and M.~Pollefeys.
\newblock {Leveraging 3D City Models for Rotation Invariant Place-of-Interest
  Recognition}.
\newblock {\em IJCV}, 2012.

\bibitem{min2018deep}
M.~Bai, G.~Mattyus, N.~Homayounfar, S.~Wang, S.~Lakshmikanth, and R.~Urtasun.
\newblock Deep multi-sensor lane detection.
\newblock In {\em IROS}, 2018.

\bibitem{bai2017deep}
M.~Bai and R.~Urtasun.
\newblock Deep watershed transform for instance segmentation.
\newblock In {\em CVPR}. IEEE, 2017.

\bibitem{Bansal14}
M.~Bansal and K.~Daniilidis.
\newblock Geometric urban geo-localization.
\newblock In {\em CVPR}, 2014.

\bibitem{deep-gil}
I.~A. B\^{a}rsan, S.~Wang, A.~Pokrovsky, and R.~Urtasun.
\newblock Learning to localize using a lidar intensity map.
\newblock In {\em CoRL}, 2018.

\bibitem{brubaker2013lost}
M.~Brubaker, A.~Geiger, and R.~Urtasun.
\newblock Lost! leveraging the crowd for probabilistic visual
  self-localization.
\newblock In {\em CVPR}, 2013.

\bibitem{fabmap}
M.~Cummins and P.~Newman.
\newblock Fab-map: Probabilistic localization and mapping in the space of
  appearance.
\newblock {\em IJRR}, 2008.

\bibitem{lsdslam}
J.~Engel, T.~Sch{\"o}ps, and D.~Cremers.
\newblock Lsd-slam: Large-scale direct monocular slam.
\newblock In {\em ECCV}, 2014.

\bibitem{Floros13}
G.~Floros, B.~van~der Zander, and B.~Leibe.
\newblock {O}pen{S}treet{SLAM}: {G}lobal vehicle localization using
  {O}pen{S}treet{M}aps.
\newblock In {\em ICRA}, 2013.

\bibitem{hartley2003multiple}
R.~Hartley and A.~Zisserman.
\newblock {\em Multiple view geometry in computer vision}.
\newblock Cambridge university press, 2003.

\bibitem{Hays08}
J.~Hays and A.~A. Efros.
\newblock im2gps: estimating geographic information from a single image.
\newblock In {\em CVPR}, 2008.

\bibitem{jaderberg2015spatial}
M.~Jaderberg, K.~Simonyan, A.~Zisserman, et~al.
\newblock Spatial transformer networks.
\newblock In {\em NIPS}, 2015.

\bibitem{kendall2015posenet}
A.~Kendall, M.~Grimes, and R.~Cipolla.
\newblock Posenet: A convolutional network for real-time 6-dof camera
  relocalization.
\newblock In {\em ICCV}, 2015.

\bibitem{kingma2014adam}
D.~P. Kingma and J.~Ba.
\newblock Adam: A method for stochastic optimization.
\newblock {\em arXiv preprint arXiv:1412.6980}, 2014.

\bibitem{kummerle2009measuring}
R.~K{\"u}mmerle, B.~Steder, C.~Dornhege, M.~Ruhnke, G.~Grisetti, C.~Stachniss,
  and A.~Kleiner.
\newblock On measuring the accuracy of slam algorithms.
\newblock {\em Autonomous Robots}, 27(4):387, 2009.

\bibitem{levinson2007map}
J.~Levinson, M.~Montemerlo, and S.~Thrun.
\newblock Map-based precision vehicle localization in urban environments.
\newblock In {\em RSS}, 2007.

\bibitem{levinson2010robust}
J.~Levinson and S.~Thrun.
\newblock Robust vehicle localization in urban environments using probabilistic
  maps.
\newblock In {\em ICRA}, 2010.

\bibitem{snavely}
Y.~Li, N.~Snavely, D.~Huttenlocher, and P.~Fua.
\newblock Worldwide pose estimation using 3d point clouds.
\newblock In {\em ECCV}, 2012.

\bibitem{linegar2015work}
C.~Linegar, W.~Churchill, and P.~Newman.
\newblock Work smart, not hard: Recalling relevant experiences for vast-scale
  but time-constrained localisation.
\newblock In {\em ICRA}, 2015.

\bibitem{hongdongli}
L.~Liu, H.~Li, and Y.~Dai.
\newblock Efficient global 2d-3d matching for camera localization in a
  large-scale 3d map.
\newblock In {\em ICCV}, 2017.

\bibitem{ma2016find}
W.-C. Ma, S.~Wang, M.~A. Brubaker, S.~Fidler, and R.~Urtasun.
\newblock Find your way by observing the sun and other semantic cues.
\newblock In {\em ICRA}, 2017.

\bibitem{matzen2013nyc3dcars}
K.~Matzen and N.~Snavely.
\newblock Nyc3dcars: A dataset of 3d vehicles in geographic context.
\newblock In {\em ICCV}, 2013.

\bibitem{moosmann2013joint}
F.~Moosmann and C.~Stiller.
\newblock Joint self-localization and tracking of generic objects in 3d range
  data.
\newblock In {\em ICRA}, 2013.

\bibitem{Mur-Artal2017}
R.~Mur-Artal and J.~D. Tardos.
\newblock {ORB-SLAM2: An Open-Source SLAM System for Monocular, Stereo, and
  RGB-D Cameras}.
\newblock {\em T-RO}, 2017.

\bibitem{nelson2015dusk}
P.~Nelson, W.~Churchill, I.~Posner, and P.~Newman.
\newblock From dusk till dawn: Localisation at night using artificial light
  sources.
\newblock In {\em ICRA}, 2015.

\bibitem{qu2015vehicle}
X.~Qu, B.~Soheilian, and N.~Paparoditis.
\newblock Vehicle localization using mono-camera and geo-referenced traffic
  signs.
\newblock In {\em IVS}. IEEE, 2015.

\bibitem{Sattler11}
T.~Sattler, B.~Leibe, and L.~Kobbelt.
\newblock Fast image-based localization using direct 2d-to-3d matching.
\newblock In {\em ICCV}, 2011.

\bibitem{schonberger2018semantic}
J.~Sch{\"o}nberger, M.~Pollefeys, A.~Geiger, and T.~Sattler.
\newblock Semantic visual localization.
\newblock {\em JPRS}, 2018.

\bibitem{schreiber2013laneloc}
M.~Schreiber, C.~Kn{\"o}ppel, and U.~Franke.
\newblock Laneloc: Lane marking based localization using highly accurate maps.
\newblock In {\em IV}, 2013.

\bibitem{jamieshotton}
J.~Shotton, B.~Glocker, C.~Zach, S.~Izadi, A.~Criminisi, and A.~Fitzgibbon.
\newblock Scene coordinate regression forests for camera relocalization in
  rgb-d images.
\newblock In {\em CVPR}, 2013.

\bibitem{wang2016torontocity}
S.~Wang, M.~Bai, G.~Mattyus, H.~Chu, W.~Luo, B.~Yang, J.~Liang, J.~Cheverie,
  S.~Fidler, and R.~Urtasun.
\newblock Torontocity: Seeing the world with a million eyes.
\newblock {\em arXiv preprint arXiv:1612.00423}, 2016.

\bibitem{WangCVPR15}
S.~Wang, S.~Fidler, and R.~Urtasun.
\newblock Holistic 3d scene understanding from a single geo-tagged image.
\newblock In {\em CVPR}, 2015.

\bibitem{WangICCV15}
S.~Wang, S.~Fidler, and R.~Urtasun.
\newblock Lost shopping! monocular localization in large indoor spaces.
\newblock In {\em ICCV}, 2015.

\bibitem{wegner2016cataloging}
J.~D. Wegner, S.~Branson, D.~Hall, K.~Schindler, and P.~Perona.
\newblock Cataloging public objects using aerial and street-level images-urban
  trees.
\newblock In {\em CVPR}, 2016.

\bibitem{welzel2015improving}
A.~Welzel, P.~Reisdorf, and G.~Wanielik.
\newblock Improving urban vehicle localization with traffic sign recognition.
\newblock In {\em ICITS}. IEEE, 2015.

\bibitem{wolcott2014visual}
R.~W. Wolcott and R.~M. Eustice.
\newblock Visual localization within lidar maps for automated urban driving.
\newblock In {\em IROS}, 2014.

\bibitem{wolcott2015fast}
R.~W. Wolcott and R.~M. Eustice.
\newblock Fast lidar localization using multiresolution gaussian mixture maps.
\newblock In {\em ICRA}, 2015.

\bibitem{yoneda2014lidar}
K.~Yoneda, H.~Tehrani, T.~Ogawa, N.~Hukuyama, and S.~Mita.
\newblock Lidar scan feature for localization with highly precise 3-d map.
\newblock In {\em IV}, 2014.

\bibitem{Zamir2016}
A.~R. Zamir, A.~Hakeem, and R.~Szeliski.
\newblock {\em {Large-Scale Visual Geo-Localization}}.
\newblock Springer, 2016.

\bibitem{jizhang}
J.~Zhang and S.~Singh.
\newblock Loam: Lidar odometry and mapping in real-time.
\newblock In {\em RSS}, 2014.

\bibitem{zhao2017pyramid}
H.~Zhao, J.~Shi, X.~Qi, X.~Wang, and J.~Jia.
\newblock Pyramid scene parsing network.
\newblock In {\em CVPR}, 2017.

\bibitem{ziegler2014video}
J.~Ziegler, H.~Lategahn, M.~Schreiber, C.~G. Keller, C.~Knoppel, J.~Hipp,
  M.~Haueis, and C.~Stiller.
\newblock Video based localization for bertha.
\newblock In {\em IV}, 2014.

\end{thebibliography}
}

\end{document}